# Paving the Way for Image Understanding:
# A New Kind of Image Decomposition is Desired

Emanuel Diamant

VIDIA-mant, P.O. Box 933, 55100 Kiriat Ono, Israel
emanl@012.net.il

**Abstract.** In this paper we present an unconventional image segmentation approach which is devised to meet the requirements of image understanding and pattern recognition tasks. Generally image understanding assumes interplay of two sub-processes: image information content discovery and image information content interpretation. Despite of its widespread use, the notion of "image information content" is still ill defined, intuitive, and ambiguous. Most often, it is used in the Shannon's sense, which means information content assessment averaged over the whole signal ensemble. Humans, however, rarely resort to such estimates. They are very effective in decomposing images into their meaningful constituents and focusing attention to the perceptually relevant image parts. We posit that following the latest findings in human attention vision studies and the concepts of Kolmogorov's complexity theory an unorthodox segmentation approach can be proposed that provides effective image decomposition to information preserving image fragments well suited for subsequent image interpretation. We provide some illustrative examples, demonstrating effectiveness of this approach.

## 1   Introduction

Meaningful image segmentation is an issue of paramount importance for image analysis and processing tasks. Natural and effortless for human beings, it is still an unattainable challenge for computer vision designers. Usually, it is approached as an interaction of two inversely directed subtasks. One is an unsupervised, bottom-up evolving process of initial image information discovery and localization. The other is a supervised, top-down propagating process, which conveys the rules and the knowledge that guide the linking and grouping of the preliminary information features into more large aggregations and sets. It is generally believed that at some higher level of the processing hierarchy this interplay culminates with the required scene decomposition (segmentation) into its meaningful constituents (objects), which then can be used for further scene analysis and interpretation (recognition) purposes.

 It is also generally believed that this way of processing mimics biological vision peculiarities, especially the characteristics of the Human Visual System (HVS). Treisman's Feature Integrating Theory [1], Biederman's Recognition-by-components theory [2], and Marr's theory of early visual information processing [3] are well known milestones of biological vision studies that for years influenced and shaped computer vision development. Although biological studies since then have seriously improved and purified their understanding of HVS properties [4], these novelties still have not find their way to modern computer vision developments.

## 2   Inherited Initial Misconceptions

The input front-end of a visual system has always been acknowledged as the most critical system's part. That is true for the biological systems and for the artificial systems as well. However, to cope with input information inundation the systems have developed very different strategies. Biological systems, in course of their natural evolution, have embraced the mechanism of Selective Attention Vision, which allows sequential part-by-part scene information gathering. Constantly moving the gaze from one scene location to another, the brain drives the eye's fovea (the eye's high-resolution sensory part) to capture the necessary information. As a result, a clear and explicit scene representation is built up and is kept in the observer's mind. (Every one, relying on his personal experience, will readily confirm this self-evident truth.)

Human-made visual systems, unfortunately, have never had such ability. Attempts (in robotic vision) to design sensors with log-polar placing of sensory elements (imitating fovea) that are attached to a steerable camera-head (imitating attentional focusing) have permanently failed. The only reasonable solution, which has survived and became the mainstream standard, was to place the photosensitive elements uniformly over the sensor's surface, covering the largest possible field of view of an imaging device. Although initially overlooked, the consequences of this move were dramatic: The bottom-up principle of input information gathering, which prescribes that every pixel in the input image must be visited and processed (normally referencing its nearest neighbors) at the very beginning, imposes an enormous system computational burden. To cope with it, unique image-processing-dedicated Digital Signal Processors (DSPs) were designed and put in duty. The latest advertised prodigy – the Analog Devices TigerSHARC – is able to provide 3,6 GFLOPS of computing power. However, despite of that, to meet real-life requirements, the use of a PCI Mezzanine Card (a BittWare product) featuring four TigerSHARCs on a single board is urgently advised. As well, applications where up to four such cards, performing simultaneously, are envisioned and afforded (delivering approximately 57 GFLOPS per cluster.)

It is worth to be mentioned here that the necessity of the DSPs usage was perfectly clear even when the "standard" image size has not exceeded 262K pixels (512x512 sensor-array). Today, as 3 –5 Megapixel arrays have became the de facto commercial standard, 16 Megapixel arrays are mastered by professionals, and 30 (up to 80) Megapixel arrays are common in military, space, medical and other quality-demanding applications, what DSP clusters arrangement would meet their bottom-up processing requirements?

The search for an answer always returns to biological vision mysteries. Indeed, the attentional vision studies have never been so widespread and extensive as in the last 5-10 years. The dynamics of eye saccadic movement is quite well understood now. As well, the rules of attention focus guidance. At the same time, various types of perceptual blindness have been unveiled and investigated. The latest research reports convincingly evidence: The hypothesis that our brain image is entire, explicit and clear – (the principal justification for the bottom-up processing) – is simply not true. It is just an illusion [5].

It will be interesting to note that despite of these impressive findings, contemporary computational models of attentional vision (and their computer vision counterparts) keep on to follow the bottom-up information gathering principle [6]. Once upon a time, as well as we are considered, someone had tried to warn about the trap of such an approach [7], but who was ready to hear? Today, a revision of the established canon is inevitable.

## 3   The Revised Segmentation Approach

Considering the results of the latest selective attention vision studies and juxtaposing them with the insights of Kolmogorov Complexity theory, which we adopt to explain the empirical biological findings, we have recently proposed a new paradigm of introductory image processing [8]. For the clarity of our discussion, we will briefly repeat some of its key points.

Taking into account the definitions of Kolmogorov's Complexity, we formulate the problem of image information content discovery and extraction as follows:

- Image information content is a set of descriptions of the observable image data structures.
- These descriptions are executable, that is, following them the meaningful part of image content can be faithfully reconstructed.
- These descriptions are hierarchical and recursive, that is, starting with a generalized and simplified description of image structure they proceed in a top-down fashion to more and more fine information details resolved at the lower description levels.
- Although the lower bound of description details is unattainable, that does not pose a problem because information content comprehension is generally fine details devoid.

An image processing strategy that can be drawn from these rules is depicted in Fig.1. As one can see, the proposed schema is comprised of three main processing paths: the bottom-up processing path, the top-down processing path and a stack where the discovered information content (the generated descriptions of it) are actually accumulated.

As it follows from the schema, the input image is initially squeezed to a small size of approximately 100 pixels. The rules of this shrinking operation are very simple and fast: four non-overlapping neighbour pixels in an image at level $L$ are averaged and the result is assigned to a pixel in a higher ($L$+1)-level image. This is known as "four children to one parent relationship". Then, at the top of the shrinking pyramid, the image is segmented, and each segmented region is labeled. Since the image size at the top is significantly reduced and since in the course of the bottom-up image squeezing a severe data averaging is attained, the image segmentation/classification procedure does not demand special computational resources. Any well-known segmentation methodology will suffice. We use our own proprietary technique that is based on a low-level (local) information content evaluation, but this is not obligatory.

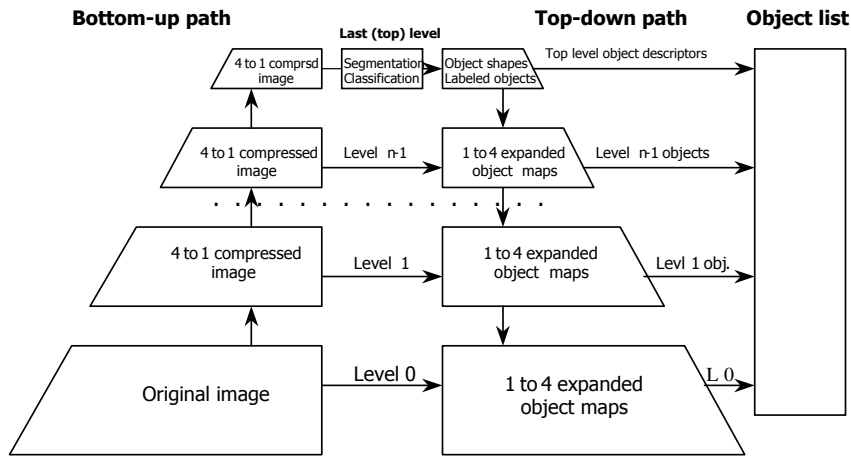

**Fig. 1.** The Schema of the proposed approach

From this point on, the top-down processing path is commenced. At each level, the two previously defined maps (average region intensity map and the associated label map) are expanded to the size of an image at the nearest lower level. Since the regions at different hierarchical levels do not exhibit significant changes in their characteristic intensity, the majority of newly assigned pixels are determined in a sufficiently correct manner. Only pixels at region borders and seeds of newly emerging regions may significantly deviate from the assigned values. Taking the corresponding current-level image as a reference (the left-side unsegmented image), these pixels can be easily detected and subjected to a refinement cycle. In such a manner, the process is subsequently repeated at all descending levels until the segmentation/classification of the original input image is successfully accomplished.

At every processing level, every image object-region (just recovered or an inherited one) is registered in the objects' appearance list, which is the third constituting part of the proposed scheme. The registered object parameters are the available simplified object's attributes, such as size, center-of-mass position, average object intensity and hierarchical and topological relationship within and between the objects ("sub-part of…", "at the left of…", etc.). They are sparse, general, and yet specific enough to capture the object's characteristic features in a variety of descriptive forms.

## 4   Illustrative Example

To illustrate the qualities of the proposed approach we have chosen a scene from the Photo-Gallery of the Natural Resources Conservation Service, USA Department of Agriculture, [9].

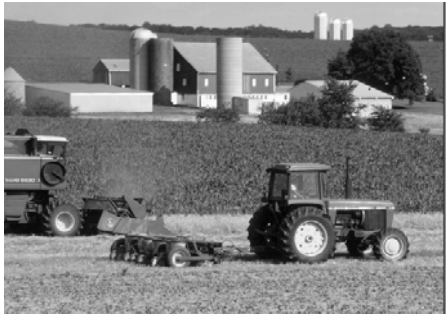
**Fig. 2.** Original image, size 1052x750 pixels

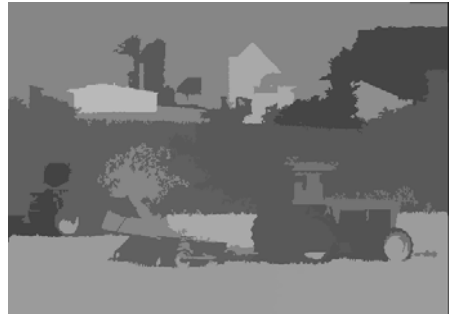
**Fig. 3.** Level 5 segmnt., 14 object-regions

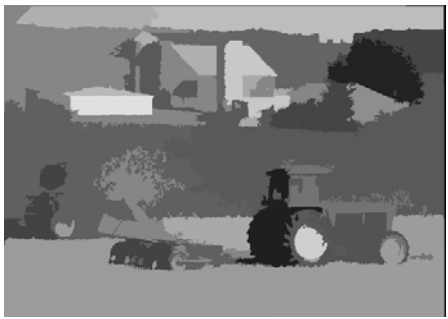
**Fig. 4.** Level 4 segmnt., 25 object-regions

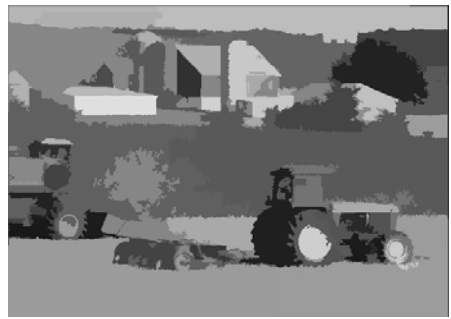
**Fig. 5.** Level 3 segmnt., 44 object-regions

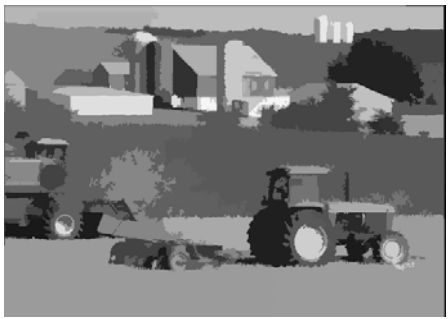
**Fig. 6.** Level 2 segmnt., 132 object-regions

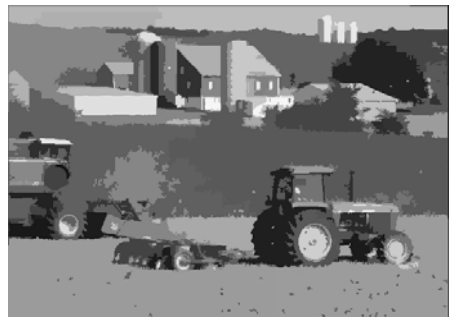
**Fig. 7.** Level 1 segmnt., 234 object-regions

Figure 2 represents the original image, Figures 3 – 7 illustrate segmentation results at various levels of the processing hierarchy. Level 5 (Fig. 3) is the topmost nearest level (For this size of the image the algorithm creates a 6-level hierarchy). Level 1 (Fig. 7) is the lower-end closest level. For space saving, we do not provide all exemplars of the segmentation succession, but for readers' convenience all presented examples are expanded to the size of the original image.

Extracted from the object list, numbers of distinguished (segmented) at each corresponding level regions (objects) are given in each figure capture.

Because real object decomposition is not known in advance, only the generalized intensity maps are presented here. But it is clear that even such simplified representations are sufficient to grasp the image concept. It is easy (for the user) now to define what region combination depicts the target object most faithfully.

## 5   Paving the Way for Image Understanding

It is clear that the proposed segmentation scheme does not produce a meaningful human-like segmentation. But it does produce the most reasonable decomposition of visually distinguishable image components, which now can be used as building blocks for an appropriate component grouping and binding procedure. Actually, image understanding arises from the correct arrangement and the right mutual coupling of the elementary information pieces gathered in the initial processing phase. The rules and the knowledge needed to execute this procedure are definitely not a part of an image. They are not an image property. They are always external, and they exist only in the head of the user, in the head of a human observer. Therefore, widespread attempts to learn them from the image stuff (automatically, unsupervised, or by supervised training) is simply a dominant misunderstanding. Nevertheless, numerous learning techniques have been devised and put in duty, including the most sophisticated biology-inspired Neural Networks. However, the trap of low-level information gathering had once again defeated the people's genuine attempts. By definition, neural network tenets assume unsupervised statistical learning, while human learning is predominantly supervised and declarative, that means, essentially natural language based and natural language supported.

What is, then, the right way to introduce to system's disposal the necessary human's knowledge and human's reasoning rules? Such a question immediately involves a subsequent challenge: how this knowledge can be or must be expressed and represented? We think that the answer is only one: as well as human's understanding relies on his world ontology, a task-specific and task-constrained ontology must be provided to system's disposal to facilitate meaningful image processing [10]. It must be human-created and domain-constrained. That means manually created by a human expert and bearing only task-specific and task-relevant knowledge about image parts concepts, their relations and interactions.

It must be specifically mentioned that these vexed questions are not only the fortune of those who are interested in image understanding issues. A huge research and development enterprise is going on now in the domain of the Semantic Web development [11]. And the unfolding of our own ideas is directly inspired by what is going on in the Semantic Web race.

Our proposed segmentation technique pretty well delineates visually discernable image parts. Essentially, the output hierarchy of segment descriptions by itself can be perceived as a form of a particular ontology, implemented in a specific description language. Therefore, to successfully accomplish the goal of knowledge incorporation, the system designer must also provide the mapping rules between these two ontologies (the mapping also has to be manually created). Because we do not intend to solve the general problem of knowledge transfer to a thinking machine, because we are always aimed on a specific and definite task, it seems that the burden of manual ontology design and its subsequent mapping can be easily carried out. If it is needed, a set of multiple ontologies can be created and cross-mapped, reflecting real life multiplicity of world to a task interaction. At least, such we hope, the things would evolve, when we shall turn to a practical realization of this idea.

## 6   Conclusions

In this paper, we have presented a new technique for unsupervised top-down-directed image segmentation, which is suitable for image understanding and content recognition applications. Contrary to traditional approaches, which rely on a bottom-up (resource exhaustive) processing and on a top-down mediating (which requires early external knowledge incorporation), our approach exploits a top-down-only processing strategy (via a hierarchy of simplified image representations). That means, considerable computational load shrinking can be attained. Especially important is its indifference to any user or task-related assumptions, its unsupervised fashion. The level of segmentation details is determined only by structures discernable in the original image data (the information content of an image, nothing other).

It must be mentioned explicitly: information content description standards like MPEG-4 and MPEG-7, which are fully relying on the concept of a recovered object, left the topic of object segmentation without the scope of the standards (for the reason of irresolvable problem's complexity). As far as we are concerned, that is the first time when a technique is proposed that autonomously yields a reasonable image decomposition (to its constituent objects), accompanied by concise object descriptions that are sufficient for reverse object reconstruction with different levels of details. Moreover, at the final image interpretation stage the system can handle entire objects, and not (as usually) pixels, from which they (obviously) are composed.